%% file: root.tex
\documentclass[letterpaper, 10 pt, conference]{ieeeconf}  

\IEEEoverridecommandlockouts                              

\usepackage{graphics} 
\usepackage{epsfig} 
\usepackage{mathptmx} 
\usepackage{times} 
\usepackage{amsmath} 
\usepackage{amssymb}  
\usepackage{graphicx}
\usepackage{cite}
\usepackage{amsfonts}
\usepackage{algorithmic}
\usepackage{graphicx}
\usepackage{textcomp}
\usepackage{multirow}
\usepackage{booktabs}
\usepackage[hyphens]{url}
\usepackage{xspace}
\usepackage[table]{xcolor}
\newcommand{\eg}{\textit{e.g.}\xspace}
\newcommand{\etal}{\textit{et al.}\xspace}

\newcommand{\ie}{\textit{i.e.}\xspace}

\usepackage[pagebackref,breaklinks,colorlinks,citecolor=blue]{hyperref}

\title{\LARGE \bf
DAP-LED: Learning Degradation-Aware Priors with CLIP for Joint Low-light Enhancement and Deblurring
}

\author{Ling Wang$^{1}$, Chen Wu$^{3}$ and Lin Wang$^{1,2,*}$
\thanks{*Corresponding author}
\thanks{$^{1}$ Ling Wang is with the AI Thrust, The Hong Kong University of Science and Technology (Guangzhou), Guangdong, China.
        {\tt\small lwang851@connect.hkust-gz.edu.cn}}%
\thanks{$^{3}$ Chen Wu is with University of Science and Technology of China, Hefei, China.
        {\tt\small wuchen5X@mail.ustc.edu.cn}}%
\thanks{$^{1,2}$Lin Wang is with AI/CMA Thrust, HKUST(GZ) and Dept. of CSE, HKUST, Hong Kong SAR, China, Email:
        {\tt\small linwang@ust.hk}}%
        }

\begin{document}

\maketitle
\thispagestyle{empty}
\pagestyle{empty}

\begin{abstract}
Autonomous vehicles and robots often struggle with reliable visual perception at night due to the low illumination and motion blur caused by the long exposure time of RGB cameras. 
Existing methods address this challenge by sequentially connecting the off-the-shelf pretrained low-light enhancement and deblurring models. Unfortunately, these methods often lead to noticeable artifacts (\eg, color distortions) in the over-exposed regions or make it hardly possible to learn the motion cues of the dark regions. 
In this paper, we interestingly find vision-language models, \eg, Contrastive Language-Image Pretraining (CLIP), can comprehensively perceive diverse degradation levels at night. 
In light of this, we propose a novel transformer-based joint learning framework, named DAP-LED, which can jointly achieve low-light enhancement and deblurring, benefiting downstream tasks, such as depth estimation, segmentation, and detection in the dark. The key insight is to leverage CLIP to adaptively learn the degradation levels from images at night. This subtly enables learning rich semantic information and visual representation for optimization of the joint tasks. 
To achieve this, we first introduce a CLIP-guided cross-fusion module to obtain multi-scale patch-wise degradation heatmaps from the image embeddings. 
Then, the heatmaps are fused via the designed CLIP-enhanced transformer blocks to retain useful degradation information for effective model optimization. 
Experimental results show that, compared to existing methods, our DAP-LED achieves state-of-the-art performance in the dark. Meanwhile, the enhanced results are demonstrated to be effective for three downstream tasks.  For demo and more results, please check the project page: \url{https://vlislab22.github.io/dap-led/}.
\end{abstract}

\section{INTRODUCTION}
\begin{figure}[t!]
    \centering
    \includegraphics[width=\linewidth]{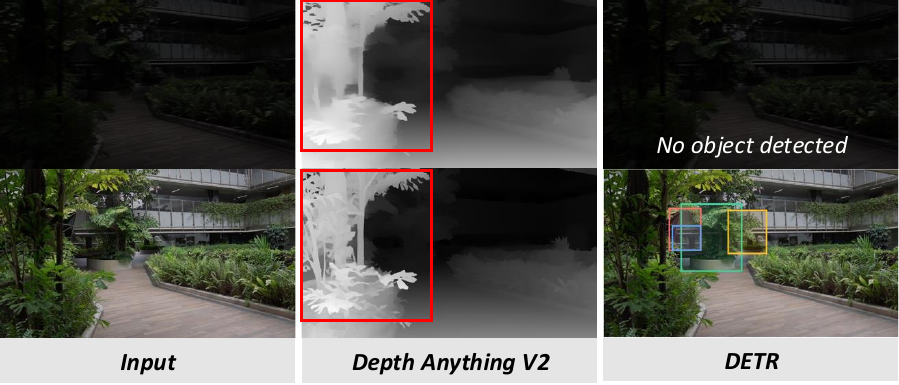}
    \vspace{-18pt}
    \caption{Comparison of depth estimation~\cite{depthanythingv2} and object detection~\cite{detr} results with night-blurred input (top) and our enhanced image as input (bottom).
    }
    \vspace{-8pt}
    \label{fig:background}
\end{figure}

Autonomous vehicles and intelligent robots are increasingly being deployed in a variety of environments, ranging from different lighting conditions. A critical challenge in these applications is to ensure robust visual perception under various adverse conditions~\cite{zhang2023perception}, particularly at night. Traditional frame-based cameras often struggle in low-light environments due to the long exposure times needed to capture sufficient illumination. This often results in significant motion blur~\cite{liu2021mba}, leading to poor visibility and reduced scene understanding accuracy.
For example, as shown in Fig.~\ref{fig:background}, without the image restoration, a dark image leads to sparse depth estimation, reduces the contrast and sharpness, and loss of depth details; More seriously, under low-light conditions, object detection is difficult, resulting in the loss of large number of targets.
In scenarios where intelligent robots are required to interact with dynamic environments, acquiring high-quality images with good visibility is crucial for safe and efficient operation. 
As a result, improving visibility in low-light conditions and addressing blurring issues are essential for vision-related tasks, especially for downstream applications like depth estimation, object detection, and semantic segmentation.

Previous approaches typically address the two tasks separately, namely, low-light enhancement~\cite{guo2020zero, li2021low} and image deblurring~\cite{chen2021blind,zhang2019deep,kupyn2019deblurgan}, each making independent assumptions about the specific problem~\cite{zhou2022lednet}. 
Therefore, simply combining these methods cannot effectively address the joint degradation caused by low light and motion blur, and may even exacerbate the degradation due to error accumulation, see Tab.~\ref{tab:lolblur}.
Specifically, methods for low light enhancement~\cite{guo2020zero, li2021low} overlook the spatial degradation caused by motion blur and would obscure clues to deblurring due to image smoothing~\cite{zhou2022lednet}. Consequently, performing low light enhancement followed by deblurring would lead to overexposure of the saturated areas of the night-blurred images, resulting in worsening blur~\cite{ye2024asp}. On the other hand, methods for deblurring~\cite{cho2021rethinking,zhang2019deep,kupyn2019deblurgan} are typically trained on datasets that contain only daytime scenes, making them unsuitable for the challenging task of directly debluring images at night due to two reasons. Firstly, the low dynamic range makes motion cues in dark regions difficult to learn. Secondly, night-blurred images often contain saturated regions~\cite{zhou2022lednet}, \eg light streaks, where the pixel values do not conform to the blur models learned from daytime data~\cite{hu2014deblurring,chen2021blind}. For these reasons, LEDNet~\cite{zhou2022lednet} takes a step forward with a unified approach to the joint task of low-light enhancement and deblurring.

\begin{figure}[t!]
    \centering
    \includegraphics[width=\linewidth]{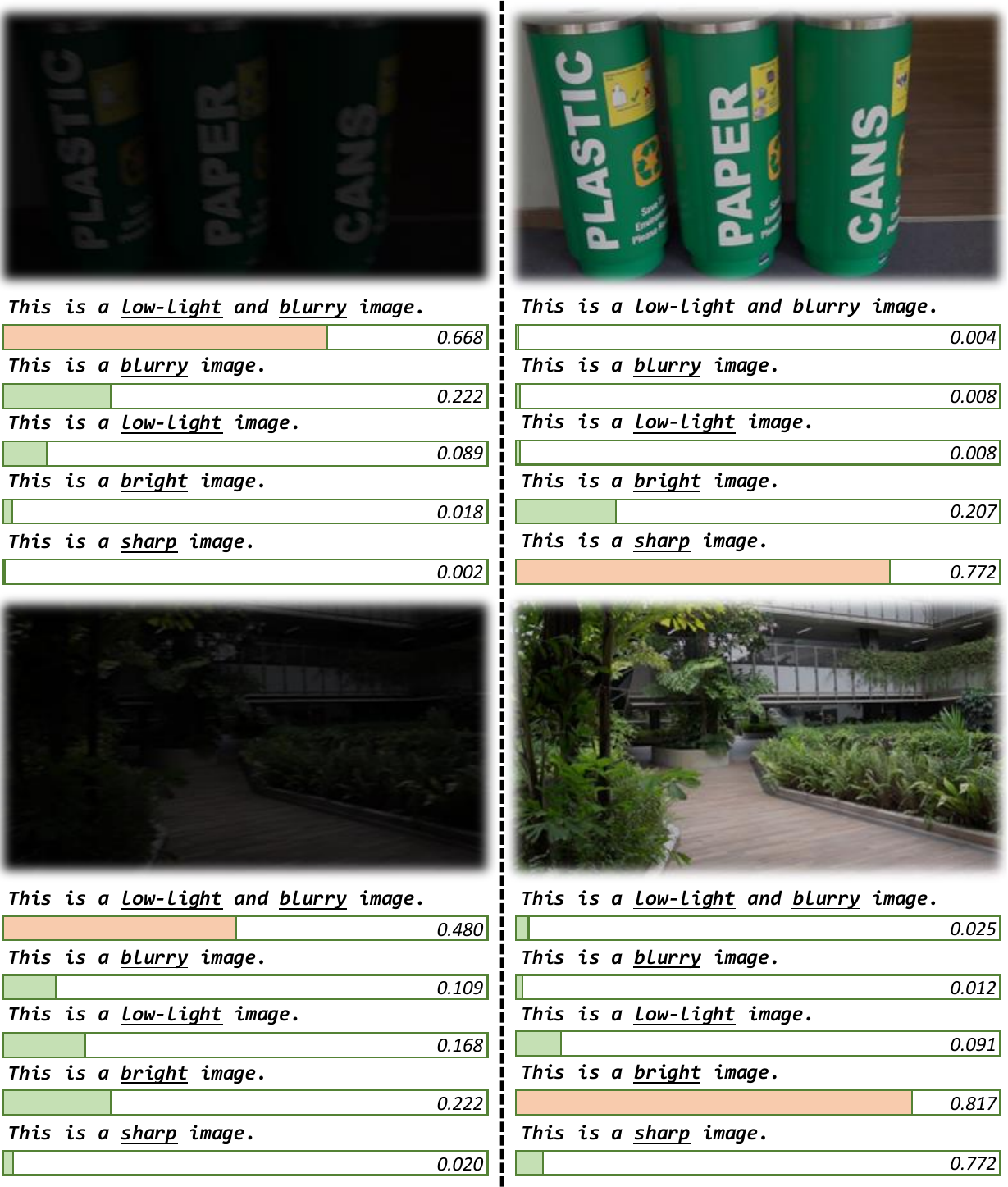}
     \vspace{-22pt}
    \caption{\textbf{Our motivation}. Here are some similarity scores between the image and text descriptions derived from the CLIP-L/14 model’s zero-shot inference. Despite some misclassifications in single degradation categories (\eg, the high score for ``bright" in a low-light image), CLIP demonstrates stronger joint degradation perception capabilities that it identifies the combined feature between low-light and blurriness in multiple images.}
    \label{fig:motivation}
    \vspace{-10pt}
\end{figure}

Building up the challenges, we observe that vision-language models, \eg, Contrastive
Language-Image Pretraining (CLIP)~\cite{radford2021learning}, offer the potential to inherently perceive and adaptively learn the joint degradation levels at night. As shown in Fig.~\ref{fig:motivation}, CLIP can inherently perceive joint degradation better than single degradation. Compared to image-based classifiers, it learns a multi-modal embedding space shared by text and image features. It contains various visual representations, including the semantic information and comprehensive degradation information. Therefore, we explore the degradation-aware priors encapsulated in CLIP for the joint task of low-light enhancement and deblurring.


In this work, we propose a novel transformer-based joint learning framework with degradation-aware CLIP priors, named \textbf{DAP-LED}, which simultaneously addresses low-light enhancement and deblurring. Our approach is beneficial for downstream tasks, such as depth estimation, segmentation, and detection in the dark. The key idea is to harness CLIP to adaptively learn degradation levels from nighttime images, thereby facilitating the extraction of rich semantic information and visual representations to optimize the joint task. To achieve this, we first introduce a CLIP-guided Cross-fusion Module (\textbf{CCM}) that generates multi-scale, patch-wise degradation heatmaps from image embeddings (see Sec.\ref{sec:clipcross}). These heatmaps are then integrated using the designed CLIP-enhanced Transformer Blocks (\textbf{CeTBs}), which preserve critical degradation information to effectively optimize the model (see Sec.\ref{sec:CETB} and Sec.~\ref{sec:cliploss}).

In summary, our main contributions are four-fold: (\textbf{I}) We leverage CLIP to inherently perceive and adaptively learn the joint degradation levels at night; (\textbf{II}) We propose a novel framework named DAP-LED that not only utilizes CLIP's strong perception ability for joint degradations but also introduces CLIP's universal visual representation for the joint task. (\textbf{III}) We introduce a CLIP-guided cross-fusion module to obtain multi-scale patch-wise degradation heatmaps and CLIP-enhanced transformer blocks to retain useful degradation information for effective model optimization. (\textbf{IV}) Extensive experiments demonstrate that our  DAP-LED, achieves remarkable image restoration effects, enefiting downstream tasks, such as depth estimation, segmentation and detection in the dark.

\section{RELATED WORK}
\noindent\textbf{Low-Light Image Enhancement (LLIE).} 
Methods based on Retinex theory are the mainstream approaches for learning-based LLIE. 
For instance, RetinexNet~\cite{wei2018deep} incorporates the Retinex theory with a Decom-Net for image decomposition and an Enhance-Net for adjusting illumination. Retinexformer~\cite{cai2023retinexformer} also integrates Retinex theory into the transformer to guide the modeling of non-local interactions among regions with varying lighting conditions. 
However, these methods often focus on brightness enhancement and denoising while neglecting the motion blur that often occurs in the robotic scene understanding process.

\noindent\textbf{Image Deblurring.} 
DeepDeblur~\cite{nah2017deep}, DMPHN~\cite{zhang2019deep}, and MPRNet~\cite{zamir2021multi} utilize a multi-stage architecture that progressively learns the degradation patterns.
By contrast, some methods, \eg Uformer~\cite{wang2022uformer} and Stripformer~\cite{tsai2022stripformer} employ local self-attention to capture long-range dependencies while maintaining low complexity. Restormer~\cite{zamir2022restormer} models global context using global channel self-attention. 
However, these methods primarily focus on deblurring tasks under normal lighting conditions, often resulting in subpar performance when applied to nighttime scene perception.

\begin{figure*}[t!]
    \centering
    \includegraphics[width=\linewidth]{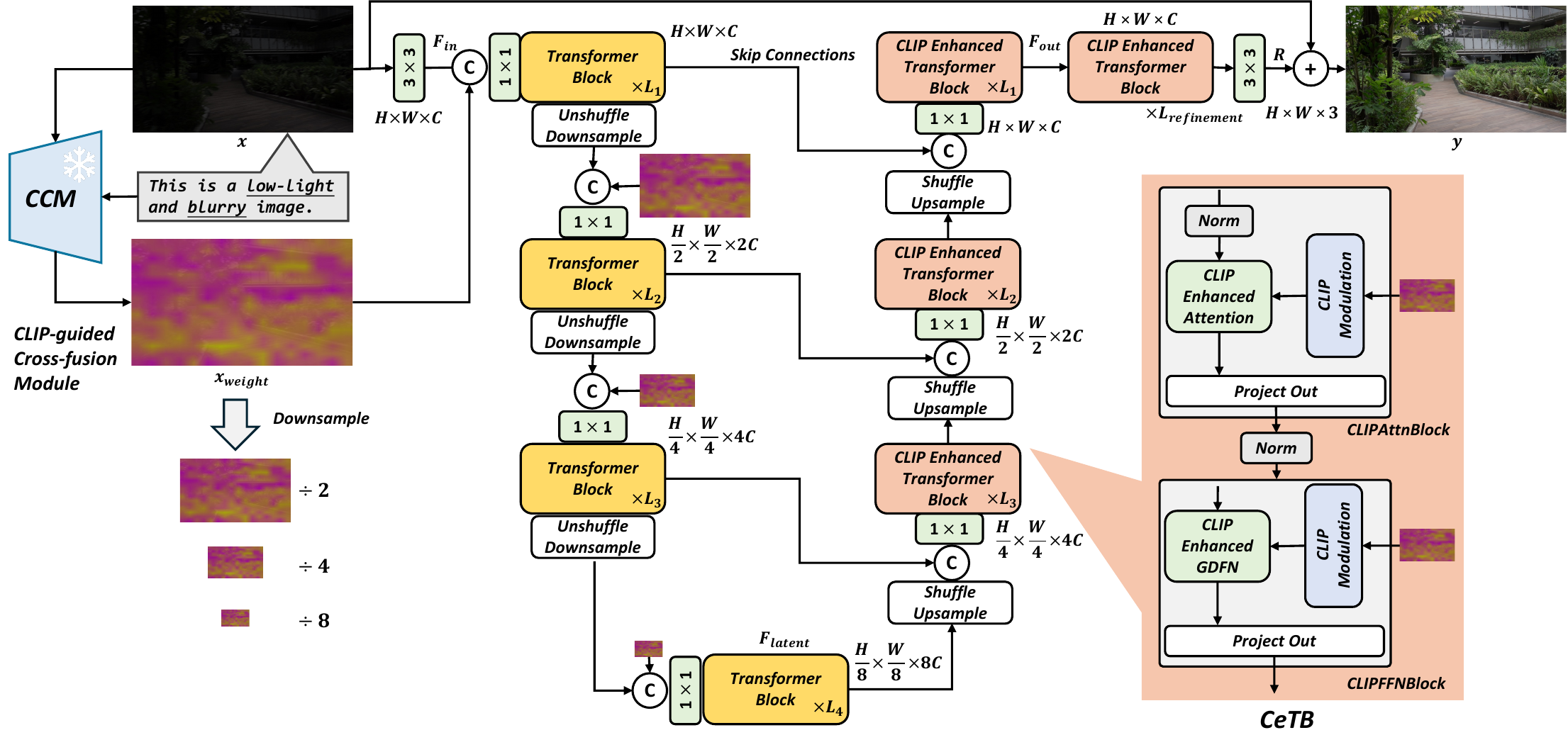}
    \vspace{-20pt}
    \caption{\textbf{Overview of our proposed DAP-LED framework}. The overall framework utilizes a 4-level transformer-based symmetric encoder-decoder structure and incorporates multiple CLIP-enhanced Transformer Blocks (\textbf{CeTBs}) at each decoding level. At the beginning, the CLIP-guided Cross-fusion Module (\textbf{CCM}) generates the multi-scale degradation-aware heatmaps, which are used at each level at encoding and decoding stages.}
    \label{fig:framework}
    \vspace{-10pt}
\end{figure*}

\noindent\textbf{Joint Low-light Enhancement and Deblurring.}
Zhou~\etal~\cite{zhou2022lednet} proposed the dataset LOL-Blur, which contains 12K low-blur/normal-sharp pairs with diverse dark and blur levels in different scenarios.
They also propose LEDNet~\cite{zhou2022lednet}, which contains an encoder for light enhancement and a decoder for deblurring. The decoder applies spatially adaptive transformation using filters generated by filter adaptive skip connection from enhanced features.
ASP-LED~\cite{ye2024asp} leverages structural priors and transformer backbones to address these issues by effectively capturing global and local features. FourierDiff~\cite{lv2024fourier} leverages Fourier priors within pre-trained diffusion models, enabling zero-shot learning that effectively handles luminance and structure degradation without needing paired data or strict degradation assumptions. Rather than relying on predefined structural assumptions or Fourier priors, by contrast, \textit{our DAP-LED framework leverages the degradation-aware CLIP priors to simultaneously address low-light enhancement and deblurring}.

\noindent\textbf{Contrastive Language-Image Pre-Training.}
Contrastive Language-Image Pre-Training, \ie CLIP, has been widely utilized for various tasks due to its outstanding cross-modal representation capabilities. These applications include areas like image manipulation~\cite{patashnik2021styleclip,wei2022hairclip,kim2022diffusionclip},  image generation~\cite{crowson2022vqgan,wang2022clip,frans2022clipdraw} and image restoration~\cite{liang2023iterative, luo2023controlling}. In this work, we leverage CLIP to adaptively learn the degradation levels from images at night. We propose the CLIP-guided cross-fusion module to obtain multi-scale patch-wise degradation heatmaps from the image embeddings. Our DAP-LED framework enables learning rich semantic information and visual representation for optimization of the joint task. 


\section{PROPOSED METHOD}
\subsection{Overall Pipeline}
The overview of our DAP-LED is depicted in Fig.~\ref{fig:framework}. Given an input image $\boldsymbol{I} \in \mathbb{R}^{H \times W \times 3}$, DAP-LED first applies a $3 \times 3$ conv layer to generate a low-level feature representation $\boldsymbol{F_{in}} \in \mathbb{R}^{H \times W \times C}$; where $H \times W$ is the spatial resolution and $C$ denotes the channel. Meanwhile, we use CLIP's image encoder and text encoder,~\ie CLIP-guided Cross-fusion Module (CCM) (Sec~\ref{sec:clipcross}), to map the image $\boldsymbol{I}$ to the CLIP embedding space and obtain the image and text cross feature embeddings $\boldsymbol{F_{weight}}$. And $\boldsymbol{F_{weight}}$ gets downsampled 3 times to correspond to the size of each level. The low-level representation $\boldsymbol{F_{in}}$ and CLIP-aware degradation weighted representation $\boldsymbol{F_{weight}}$ is then processed through a 4-level UNet-like hierarchical encoder-decoder, producing deep features $\boldsymbol{F}_{latent} \in \mathbb{R}^{\frac{H}{8} \times \frac{W}{8} \times 8C}$. The encoder consists of multiple transformer blocks, and the decoder consists of CLIP-enhanced Transformer Blocks (CeTBs) (Sec~\ref{sec:CETB}). Finally, DAP-LED employs a conv layer to generate the residual image $\boldsymbol{R} \in \mathbb{R}^{H \times W \times 3}$ from $\boldsymbol{F}_{latent}$. The restored image is then obtained by adding the degraded image to the residual, \textit{i.e.}, $\boldsymbol{\hat{I}} = \boldsymbol{I} + \boldsymbol{R}$. Now, let's describe the details of CCM and CeTBs.

\begin{figure}[t!]
    \centering
    \includegraphics[width=0.8\linewidth]{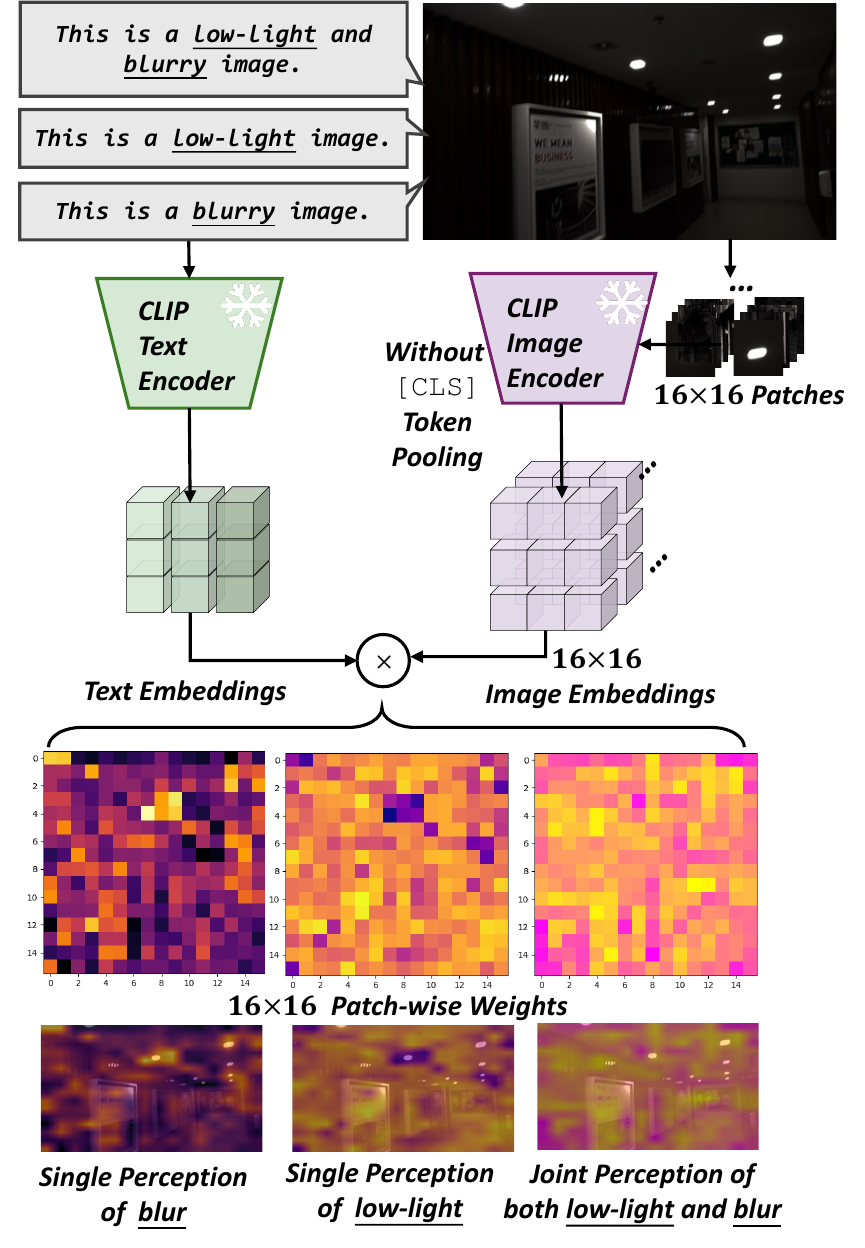}
     \vspace{-13pt}
    \caption{\textbf{Overview of CCM}. Here is a visualization of the patch-wise similarity weights between the image and text embeddings. The heatmap on the bottom right shows the model’s joint perception of low-light and blur, as well as single perceptions of low-light and blur individually.}
    \label{fig:heatmap}
    \vspace{-10pt}
\end{figure}

\begin{figure*}[t!]
    \centering
    \includegraphics[width=\linewidth]{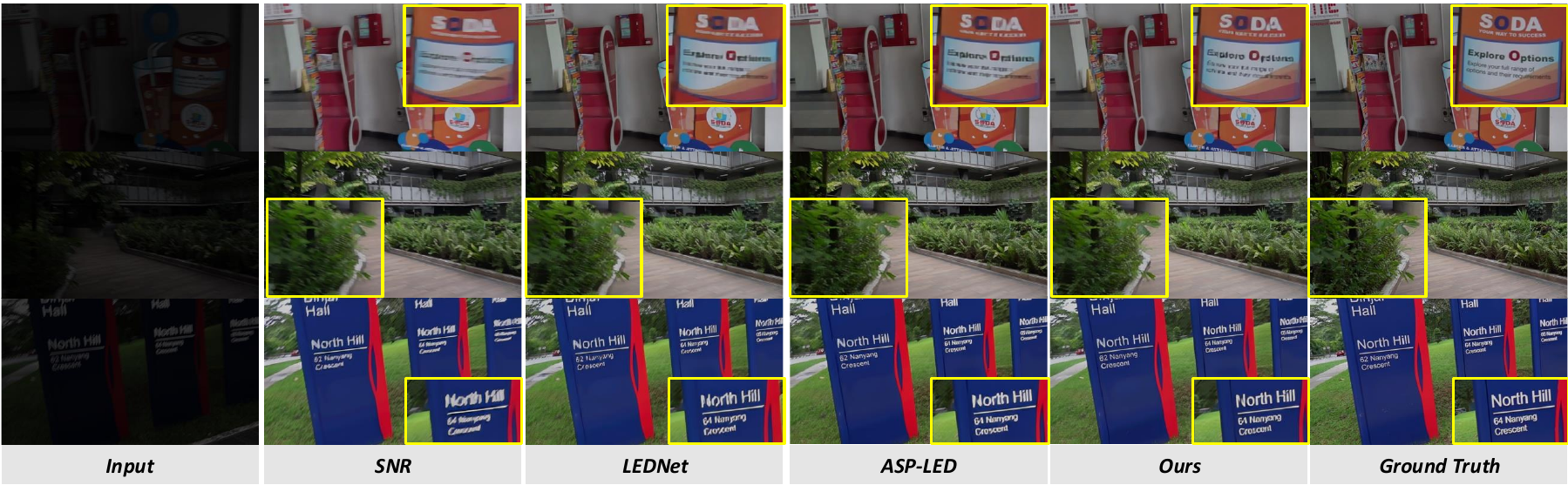}
     \vspace{-18pt}
    \caption{Visual quality comparison on the LOL-Blur dataset. Please check the zoom-in patches to observe the details.}
    \label{fig:lolblur}
\end{figure*}

\subsection{CLIP-guided Cross-fusion Module (CCM)}
\label{sec:clipcross}
The CLIP-guided Cross-fusion Module (CCM) plays a pivotal role in leveraging CLIP to jointly perceive and learn the low-light and blur characteristics of images. As shown in Fig.~\ref{fig:heatmap}, the module begins by passing the input image $\boldsymbol{I}$ through the CLIP image encoder $\operatorname{\mathbf{E}_{i}}$, where it is split into $16 \times 16$ patch-wise image tokens without using the [CLS] token. Simultaneously, a set of predefined text prompts, such as ``\textit{This is a low-light and blurry image}." are processed through the CLIP text encoder $\operatorname{\mathbf{E}_{t}}$, which generates text embeddings representing these specific degradations. 
Next, the image embeddings and text embeddings are fused to compute patch-wise similarity weights. These weights are utilized to generate degradation-level heatmaps, which visually represent the contribution of each patch to the overall perception of joint degradation. The heatmap highlights the spatial regions of the image that are most affected by degradations, allowing the model to focus on relevant areas for adaptive enhancement. This process is summarized as:
\begin{equation}
\boldsymbol{F}_{weight}=\operatorname{\boldsymbol{F}_{in}}\odot[\operatorname{\mathbf{E}_{i}}({\boldsymbol{F}_{in}})*\operatorname{\mathbf{E}_{t}}(prompt)], \\
\end{equation}
where $\boldsymbol{F}_{weight}$ is intermediate results of clip enchanced image, $\odot$ denotes patch-wise multiplication, $*$ denotes matrix multiplication, $\mathbf{E}_{i}$ is CLIP image encoder without pooling, $\mathbf{E}_{t}$ is CLIP text encoder.

\subsection{CLIP-Enhanced Transformer Blocks (CeTBs)}
\label{sec:CETB}
As shown in Fig.~\ref{fig:framework}, these blocks are designed to facilitate multi-scale joint-degradation feature learning in that each type of image degradation exhibits distinct perturbed patterns~\cite{wang2022uformer}. Previous studies~\cite{karras2020analyzing,wang2022uformer} have shown that incorporating multi-scale noise components into feature maps can significantly improve a model’s ability to handle diverse disturbances, enhancing the recovery of detailed information from degraded images. However, using uncontrollable noise terms in image restoration tasks can be problematic, as it introduces randomness, making it difficult to regulate the finer details of the recovered image~\cite{saharia2022image}.

To address this issue, we integrate degradation-aware CLIP priors at each level of the decoder, analogous to controllable noise elements. The CeTB is based on the Transformer block~\cite{zamir2022restormer} but with a key distinction: incorporating joint degradation-aware CLIP priors within the attention mechanism. This results in an updated formulation of the CLIP Enhanced Attention:
\begin{align}
&\hat{\boldsymbol{F}}_{in}=\boldsymbol{F}_{in}+\operatorname{Modulation}(\boldsymbol{F}_c),\\
&\boldsymbol{F}_{out}=\operatorname{Attn}(W_p^Q\hat{\boldsymbol{F}}_{in},W_p^K\hat{\boldsymbol{F}}_{in},W_p^V\hat{\boldsymbol{F}}_{in})+\boldsymbol{F}_{in},
\end{align}
where $W_p^Q$, $W_p^K$, and $W_p^V$ are projection matrices, and $\operatorname{Attn}(\cdot)$ denotes the attention mechanism. By leveraging degradation-aware CLIP priors, the CeTB enhances the model’s ability to handle joint degradation patterns across different scales while maintaining control over the restoration process.
\subsection{Optimization}
\label{sec:cliploss}
To eliminate the joint-degradation in the observed image, we utilize the Charbonnier loss as the basic reconstruction loss $\mathbf{L}_{rec}$ in the RGB color space, which can be denoted as:
\begin{equation}
\mathbf{L}_{rec}=\sqrt{\left\|\hat{\boldsymbol{I}}-\boldsymbol{I}\right\|^2+\epsilon^2},
\end{equation}
where $\mathbf{I}$ and $\hat{\mathbf{I}}$ represent the restored image and ground truth, respectively. $\epsilon$ is a constant. We also introduce the CLIP-aware loss to guide the restored result. The CLIP-aware loss comprises two components: the identity loss $\mathbf{L}_{identity}$ and the CLIP loss $\mathbf{L}_{clip}$. By integrating the degradation-aware $\mathbf{L}_{clip}$, Our method promotes the restored image to align closely with the ground truth in the CLIP embedding space, ensuring the preservation of semantic consistency. Mathematically,
\begin{align}
\mathbf{L}_{identity}&=\left\|\operatorname{E_i}(\hat{\boldsymbol{I}})-\operatorname{E_i}(\boldsymbol{I})\right\|_2,\\
\mathbf{L}_{clip}&=\frac{\operatorname{E_i}(\hat{\boldsymbol{I}})\cdot \operatorname{E_t}(prompt)}{|\operatorname{E_i}(\hat{\boldsymbol{I}})||\operatorname{E_t}(prompt)|},
\end{align}
where $\operatorname{E_i}(\cdot)$ and $\operatorname{E_t}(\cdot)$ denote the image and text encoder of CLIP, respectively. $prompt$ represents a concise description of the joint-degradation depicted in the input image. Finally, the overall loss function can be expressed as:
\begin{equation}
\mathbf{L}_{total}=\mathbf{L}_{rec}+\lambda_{identity}\mathbf{L}_{identity}+\lambda_{clip}\mathbf{L}_{clip},
\end{equation}
where $\lambda_{identity}$ and $\lambda_{clip}$ are the trade-off parameters.
\input{table/lolblur}

\section{EXPERIMENTS}
\subsection{Implementation Details and Datasets} 
\begin{figure*}[t!]
    \centering
    \includegraphics[width=\linewidth]{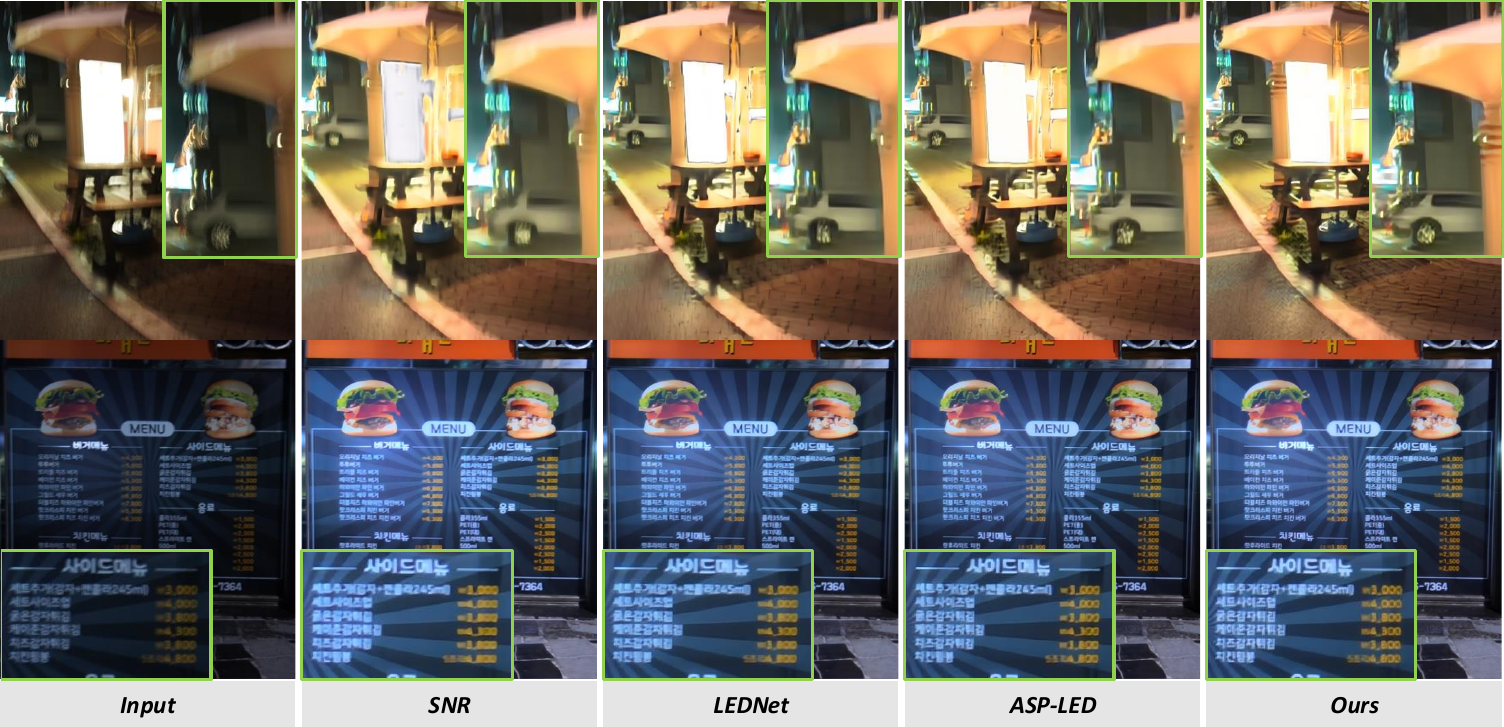}
    \vspace{-18pt}
    \caption{Visual quality comparison on the Real-LOL-Blur dataset. Please check the zoom-in patches to observe the details.}
    \label{fig:reallolblur}
\end{figure*}
\noindent\textbf{Implementation.} We conduct experiments in PyTorch on 4 NVIDIA A800 GPUs. To optimize the network, we employ the Adam optimizer with a learning rate $1\times10^{-4}$. The network is trained for 500k iterations with a batch size of 4. During training, we utilize cropped patches of size $512\times512$ as input, and to augment the training data, random horizontal and vertical flips are applied to the input images. The architecture of our method consists of a 4-level encoder-decoder, with varying numbers of Transformer blocks at each level, specifically $[1, 2, 2, 4]$ from level-1 to level-4. For CLIP, we use the ViT-L/14 model to extract feature embeddings. Additionally, hyperparameters are empirically set as: $\lambda_{identity}=0.1$ and $\lambda_{clip}=0.01$.

\noindent\textbf{Benchmark Datasets.} We utilize the LOL-Blur dataset~\cite{zhou2022lednet} to train our model and assess its generalization capabilities using the real-world Real-LOL-Blur dataset. For details of these datasets, refer to our project homepage.
For evaluating the performance on the LOL-Blur dataset, we use PSNR, SSIM, and LPIPS. In contrast, for Real-LOL-Blur, we rely on perceptual metrics such as MUSIQ, NRQM, and NIQE.

\subsection{Results on Synthetic Datasets} 
In the experiments of the LOL-Blur dataset, we compare our method with some state-of-the-art approaches, including two cascade low-light enhancement and deblurring methods Zero-DCE~\cite{guo2020zero} $\rightarrow$ MIMO~\cite{cho2021rethinking}, RUAS~\cite{liu2021retinex} $\rightarrow$ MIMO~\cite{cho2021rethinking}, three cascade deblurring and low-light enhancement methods Chen~\cite{chen2021blind} $\rightarrow$ Zero-DCE~\cite{guo2020zero}, DeblurGAN-v2~\cite{kupyn2019deblurgan} $\rightarrow$ Zero-DCE~\cite{guo2020zero}, MIMO~\cite{cho2021rethinking} $\rightarrow$ Zero-DCE~\cite{guo2020zero}, three low-light enhancement models KinD++~\cite{zhang2021beyond}, DRBN~\cite{yang2021band}, SNR~\cite{xu2022snr}, three deblurring models DeblurGANv2~\cite{kupyn2019deblurgan}, DMPHN~\cite{zhang2019deep}, MIMO~\cite{cho2021rethinking}, and the two joint low-light enhancement and deblurring methods LEDNet~\cite{zhou2022lednet} and ASP-LED~\cite{ye2024asp}. 
As illustrated in Tab.~\ref{tab:lolblur}, our method achieved state-of-the-art performance for this task, surpassing other models in terms of PSNR, SSIM, and LPIPS metrics. The visual comparisons in Fig.~\ref{fig:lolblur} further demonstrate that our approach excels at restoring curved edge details and retrieving finer, sharper textural features when compared to competing methods. These results confirm that our model performs not only well quantitatively but also qualitatively in enhancing images under challenging low-light and blurry conditions, demonstrating superior recovery of intricate details that are vital for downstream applications.

\subsection{Results on the Real-World Dataset} 
\input{table/reallolblur}
Tab.~\ref{tab:lolblurreal} presents the quantitative evaluation of different methods on the Real-LOL-Blur dataset using the MUSIQ, NRQM, and NIQE metrics. Our proposed DAP-LED method achieves the best performance across the MUSIQ and NRQM scores. Although our NIQE score is slightly higher than ASP-LED’s, it remains highly competitive, further confirming the effectiveness of our model in restoring image quality under challenging low-light and blur conditions. Fig.~\ref{fig:reallolblur} compares the restored images across different methods, further demonstrating the superiority of our approach. In particular, our model recovers more details and improves the clarity of both the environment and the textual information on the signage, as shown in the magnified regions. Compared to the input and other methods like SNR, LEDNet and ASP-LED, our results offer better contrast and sharpness, eliminating much of the blurring and enhancing the overall visibility in low-light scenarios. This visual quality comparison further confirms the strong quantitative results, highlighting the effectiveness of our method in practical scenarios.

\input{table/abla1}
\input{table/abla2}

\subsection{Ablation Study} 
In Tab.~\ref{table:abla1} and Tab.~\ref{table:abla2}, we present the results of the ablation studies performed on DAP-LED to evaluate the impact of different prompt configurations and loss functions on image restoration performance. Tab.~\ref{table:abla1} compares the effectiveness of using different prompt learning strategies, including deblur prompt, lowlight prompt, and their combination (Joint Prompt). The results indicate that while both individual prompts improve image quality, the joint prompt yields the highest PSNR and SSIM, demonstrating the benefits of leveraging both aspects simultaneously.

Tab.~\ref{table:abla2} presents an ablation study on the losses used during training, showing the impact of adding identity loss (\(\mathbf{L}_{identity}\)) and CLIP loss (\(\mathbf{L}_{clip}\)) alongside the reconstruction loss (\(\mathbf{L}_{rec}\)). The results show a progressive improvement in PSNR and SSIM as these additional losses are included. The highest performance is achieved with the full loss combination, highlighting the importance of incorporating both identity and CLIP constraints for improved image restoration.

\begin{figure}[t!]
    \centering
    \includegraphics[width=\linewidth]{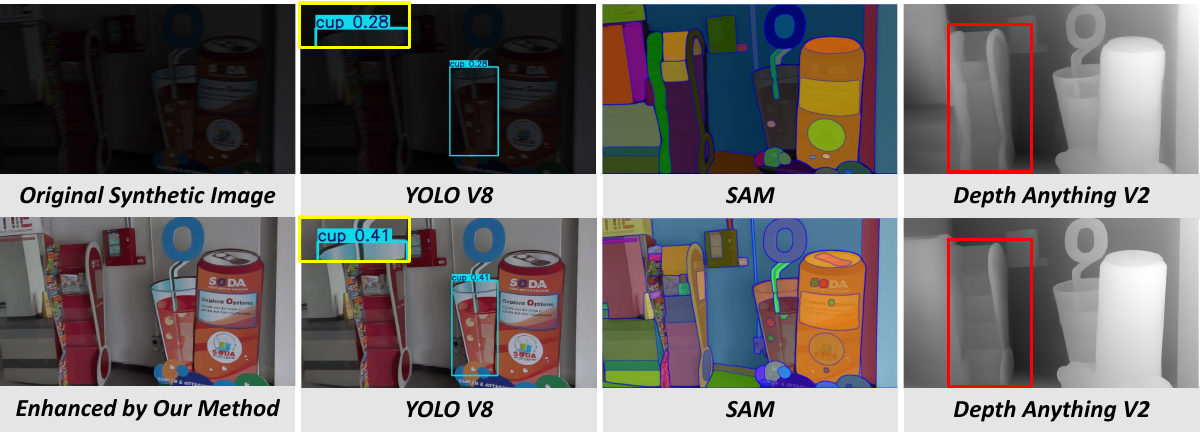}
     \vspace{-16pt}
    \caption{Visual Results of YOLO V8, SAM and Depth Anything V2 on the \textbf{synthetic} LOL-Blur datasset.}
    \label{fig:downstream_syn}
\end{figure}

\begin{figure}[t!]
    \centering
    \includegraphics[width=\linewidth]{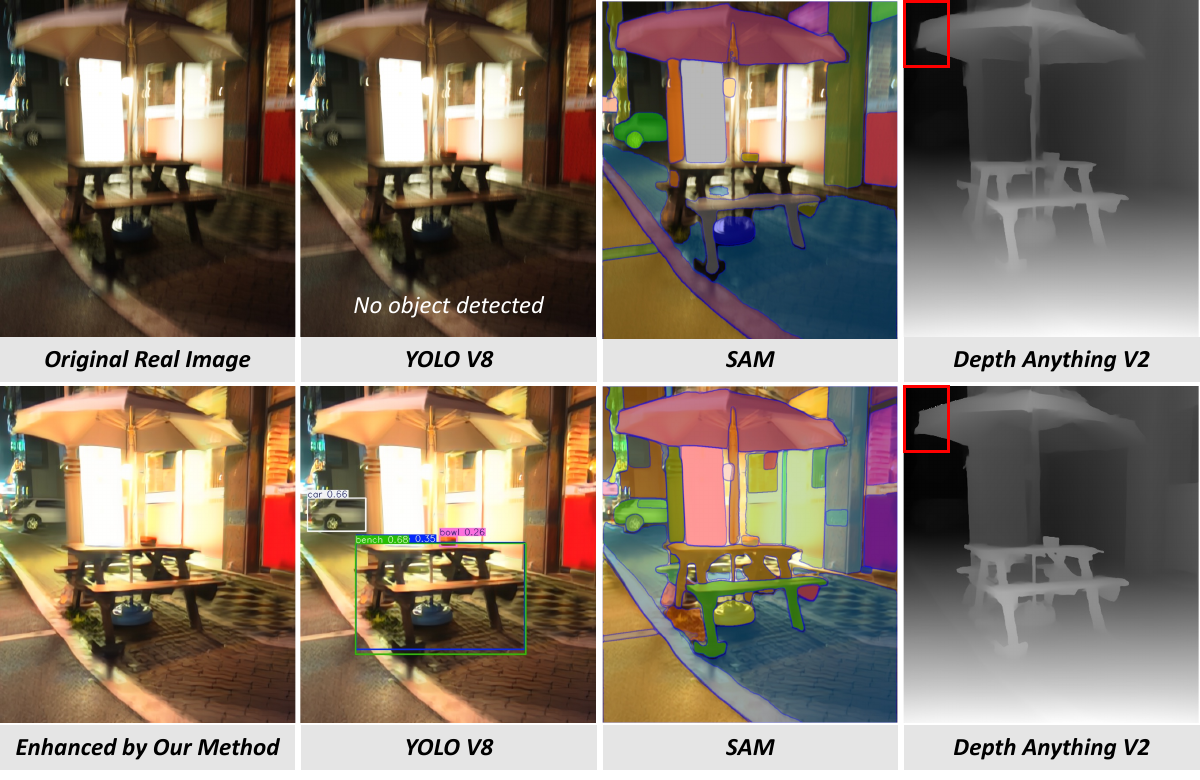}
        \vspace{-16pt}
    \caption{Visual Results of YOLO V8, SAM and Depth Anything V2 on the Real LOL-Blur dataset.}
    \label{fig:downstream_real}
    \vspace{-10pt}
\end{figure}

\subsection{Applications to Downstream Vision Tasks}
To verify the effectiveness of our DAP-LED framework, we further apply the enhanced results to three downstream vision tasks, depth estimation (with DepthAnything V2~\cite{depth_anything_v2}), segmentation (SAM~\cite{kirillov2023segment}), and detection (YoloV8~\cite{YOLOv8}). We provide comparisons of three downstream tasks for the night-blurred image and enhanced image. Fig.~\ref{fig:downstream_syn} and Fig.~\ref{fig:downstream_real} depict the qualitative results for 1) the synthetic LOL-Blur dataset and 2) the real-world Real-LOL-Blur dataset, respectively. The bottom row of each figure demonstrates better results with our enhanced image as input. YOLO V8 enhances both the precision of object detection and the total number of detected objects. SAM delivers clearer segmentation boundaries, while Depth Anything V2 generates more detailed and accurate depth maps.

\section{CONCLUSIONS}
In this paper, we introduced DAP-LED, a novel joint low-light enhancement and deblurring framework that leverages the powerful multi-modal perception capabilities of the CLIP model. By utilizing CLIP’s text-image embedding space, our method can perceive and restore complex image degradations more effectively. Through extensive experiments, we demonstrated that DAP-LED achieves state-of-the-art performance on real night-blurred images, beneficial for three downstream tasks in the dark for robotic scene understanding. 
Future work will focus on expanding this framework to handle more diverse degradations and further optimizing the restoration process for real-time applications.

\bibliographystyle{IEEEtran}
\bibliography{IEEEfull}
\end{document}

%% file: table/lolblur.tex
\begin{table}[t!]
\caption{\textbf{Quantitative evaluation on LOL-Blur dataset.} PSNR/SSIM$\uparrow$: the higher, the better; LPIPS$\uparrow$: the lower, the better. Best values are indicated with \textbf{bold} text.}
 \vspace{-9pt}
\label{tab:lolblur}
\resizebox{\linewidth}{!}{
\begin{tabular}{cccc}
\toprule[0.3mm]
\multicolumn{1}{c|}{Methods}                             & $PSNR \uparrow$ & \multicolumn{1}{c|}{$SSIM \uparrow$} & $LPIPS \downarrow$ \\ 
\bottomrule[0.3mm]
\multicolumn{4}{c}{\cellcolor{olive!10}Enhancement $\rightarrow$ Deblurring}\\ 
\toprule[0.3mm]
\multicolumn{1}{c|}{Zero-DCE $\rightarrow$ MIMO} & 17.68 & \multicolumn{1}{c|}{0.542} & 0.510 \\
\multicolumn{1}{c|}{RUAS $\rightarrow$ MIMO}     & 17.81 & \multicolumn{1}{c|}{0.569} & 0.523 \\ \bottomrule[0.3mm]
\multicolumn{4}{c}{\cellcolor{lime!10}Deblurring $\rightarrow$ Enhancement}                                      \\ \toprule[0.3mm]
\multicolumn{1}{c|}{Chen~\etal $\rightarrow$ Zero-DCE} & 17.02 & \multicolumn{1}{c|}{0.502} & 0.516 \\
\multicolumn{1}{c|}{DeblurGAN-v2 $\rightarrow$ Zero-DCE} & 18.33           & \multicolumn{1}{c|}{0.589}           & 0.476              \\
\multicolumn{1}{c|}{MIMO $\rightarrow$ Zero-DCE} & 17.52 & \multicolumn{1}{c|}{0.57}  & 0.498 \\ \bottomrule[0.3mm]
\multicolumn{4}{c}{\cellcolor{green!10}Training on LOL-Blur}\\ 
\toprule[0.3mm]
\multicolumn{1}{c|}{KinD++}                      & 21.26 & \multicolumn{1}{c|}{0.753} & 0.359 \\
\multicolumn{1}{c|}{DRBN}                        & 21.78 & \multicolumn{1}{c|}{0.768} & 0.325 \\
\multicolumn{1}{c|}{DeblurGAN-v2}                & 22.30 & \multicolumn{1}{c|}{0.745} & 0.356 \\
\multicolumn{1}{c|}{DMPHN}                       & 22.20 & \multicolumn{1}{c|}{0.817} & 0.301 \\
\multicolumn{1}{c|}{MIMO}                        & 22.41 & \multicolumn{1}{c|}{0.835} & 0.262 \\ 
\multicolumn{1}{c|}{LEDNet}                      & 25.74 & \multicolumn{1}{c|}{0.850} & \underline{0.224} \\ 
\multicolumn{1}{c|}{ASP-LED}                     & \textbf{26.73} & \multicolumn{1}{c|}{\textbf{0.866}} & \textbf{0.199} \\ 
\hline
\multicolumn{1}{c|}{Ours}                        & \underline{26.42} & \multicolumn{1}{c|}{\underline{0.853}} & 0.234 \\ 
\bottomrule[0.3mm]
\end{tabular}}
\vspace{-10pt}
\end{table}

%% file: table/reallolblur.tex
\begin{table}[t!]
\caption{\textbf{Quantitative evaluation on Real-LOL-Blur dataset.} MUSIQ/NRQM$\uparrow$: the higher, the better; NIQE$\uparrow$: the lower, the better. Best values are indicated with \textbf{bold} text.}
\vspace{-8pt}
\label{tab:lolblurreal}
\resizebox{.88\linewidth}{!}{
\begin{tabular}{cccc}
\toprule[0.3mm]
\multicolumn{1}{c|}{Method} & $MUSIQ \uparrow$ & \multicolumn{1}{c|}{$NRQM \uparrow$} & $NIQE \downarrow$ \\ 
\bottomrule[0.3mm]
\multicolumn{4}{c}{\cellcolor{olive!10}Enhancement $\rightarrow$ Deblurring}             \\
\toprule[0.3mm]
\multicolumn{1}{c|}{Zero-DCE → MIMO} & 39.36 & \multicolumn{1}{c|}{5.206} & 4.459 \\
\multicolumn{1}{c|}{RUAS → MIMO}     & 34.39 & \multicolumn{1}{c|}{3.322} & 6.812 \\
\bottomrule[0.3mm]
\multicolumn{4}{c}{\cellcolor{lime!10}Deblurring $\rightarrow$ Enhancement}              \\
\toprule[0.3mm]
\multicolumn{1}{c|}{Chen~\etal → Zero-DCE} & 45.79 & \multicolumn{1}{c|}{5.844} & 5.043 \\
\multicolumn{1}{c|}{MIMO → Zero-DCE} & 28.36 & \multicolumn{1}{c|}{3.697} & 6.892 \\
\bottomrule[0.3mm]
\multicolumn{4}{c}{\cellcolor{green!10}Training on LOL-Blur}     \\
\toprule[0.3mm]
\multicolumn{1}{c|}{KinD++}          & 31.74 & \multicolumn{1}{c|}{3.854} & 7.299 \\
\multicolumn{1}{c|}{DRBN}            & 31.27 & \multicolumn{1}{c|}{4.019} & 7.129 \\
\multicolumn{1}{c|}{SNR}             & 34.58 & \multicolumn{1}{c|}{4.662} & 5.31  \\
\multicolumn{1}{c|}{DMPHN}           & 35.08 & \multicolumn{1}{c|}{4.47}  & 5.19  \\
\multicolumn{1}{c|}{MIMO}            & 35.37 & \multicolumn{1}{c|}{5.14}  & 4.851 \\
\multicolumn{1}{c|}{LEDNet}          & 39.11 & \multicolumn{1}{c|}{\textbf{5.643}} & 4.764 \\
\multicolumn{1}{c|}{ASP-LED}         & \underline{46.95} & \multicolumn{1}{c|}{5.583} & \underline{4.442} \\ \hline
\multicolumn{1}{c|}{Ours}         & \textbf{47.21} & \multicolumn{1}{c|}{\underline{5.612}} &  \textbf{4.323}\\
\bottomrule[0.3mm]
\end{tabular}}
\vspace{-5pt}
\end{table}

%% file: table/abla1.tex
\begin{table}[t!]
\centering
\caption{Ablation study of the prompt used in DAP-LED.}
\vspace{-10pt}
\label{table:abla1}
\resizebox{.8\linewidth}{!}{
\begin{tabular}{l|l|l}
\bottomrule[0.3mm]
Experiment on Prompt Learning  & $PSNR \uparrow$  & $SSIM \uparrow$  \\ \hline
(1) deblur prompt              & 23.12 & 0.824 \\
(2) lowlight prompt            & 24.45 & 0.842 \\\hline
(3) joint prompt               & 26.42 & 0.853 \\ 
\bottomrule[0.3mm]
\end{tabular}}
\vspace{-5pt}
\end{table}

%% file: table/abla2.tex
\begin{table}[t!]
\centering
\caption{Ablation study of the losses of DAP-LED.}
\vspace{-10pt}
\label{table:abla2}
\resizebox{.9\linewidth}{!}{
\begin{tabular}{l|l|l}
\bottomrule[0.3mm]
Experiment on Losses  & $PSNR \uparrow$  & $SSIM \uparrow$  \\ \hline
(1) $\mathbf{L}_{rec}$   & 25.73 & 0.839 \\
(2) $\mathbf{L}_{rec}+\lambda_{identity}\mathbf{L}_{identity}$     & 26.31 & 0.847 \\\hline
(3) $\mathbf{L}_{rec}+\lambda_{identity}\mathbf{L}_{identity}+\lambda_{clip}\mathbf{L}_{clip}$  & 26.42 & 0.853 \\ 
\bottomrule[0.3mm]
\end{tabular}}
\vspace{-10pt}
\end{table}